# Flexible and slim device switching air blowing and suction by a single airflow control

Seita Nojiri[1], *Student member, IEEE*, Toshihiro Nishimura[2], Kenjiro Tadakuma[3], *Member, IEEE,* and Tetsuyou Watanabe[2], *Member, IEEE*

*Abstract*—This study proposes a soft robotic device with a slim and flexible body that switches between air blowing and suction with a single airflow control. Suction is achieved by jet flow entraining surrounding air, and blowing is achieved by blocking and reversing jet flow. The thin and flexible flap gate enables the switching. Air flow is blocked while the gate is closed and passes through while the gate is open. The opening and closing of the flap gate are controlled by the expansion of the inflatable chambers installed near the gate. The extent of expansion is determined by the upstream static pressure. Therefore, the gate can be controlled by the input airflow rate. The dimensions of the flap gate are introduced as a design parameter, and we show that the parameter contributes to the blowing and suction capacities. We also experimentally demonstrate that the proposed device is available for a variable friction system and an end effector for picking up a thin object covered with dust.

## I. INTRODUCTION

A soft robot driven by a fluid has a simple structure made of flexible materials. The simplest structure has only inflatable flow paths in the body. Body deformation is caused by fluid flow that is added to or removed from the flow paths and enables various functions, such as moving, grasping, and energy storage [1]. Flow control is typically achieved using solenoid valves [2]–[5]. The number of solenoid valves required increases with the number of flow switches, increasing the complexity, mechanical structure, and cost of the control system [6]. To address this issue, different or multiple actuating functions performed by 1-degree-of-freedom (1-DOF) fluid flow controls have attracted considerable attention. Typical examples are [7] and [8], in which switching between different actuations is realized by controlling the input air pressure or flow rate. However, unresolved challenges remain. One of these is the reversal of the flow direction. The fluid source provides either discharge or suction. Another challenge is that the devices must be slim and flexible.

Thus, this study develops a flow-direction reversal (FDR) device (Fig. 1) that switches between blowing and suction and

Manuscript received: November 18, 2022; Revised: February 6, 2023; Accepted: February 28, 2023. This letter was recommended for publication by Associate Editor and Editor Yong-Lae Park upon evaluation of the reviewers' comments. *(Corresponding authors: Tetsuyou Watanabe; Toshihiro Nishimura.)*

This work was supported by JST Moonshot R&D Grant Number JPMJMS2034, MEXT/JSPS KAKENHI Grant Number JP21H01286, JP21K19790, and JST, the establishment of university fellowships toward the creation of science technology innovation, Grant Number JPMJFS2116.

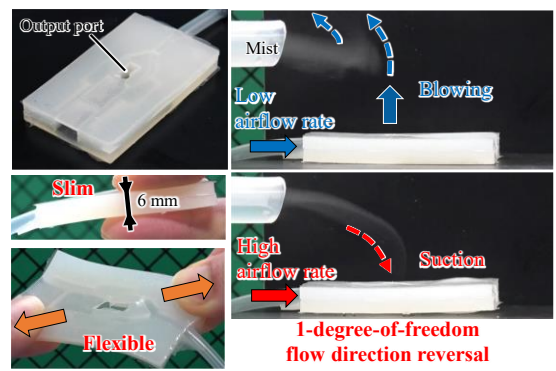

**Fig. 1.** Flow-direction reversal (FDR) device (The input airflow rate was 10 and 30 L/min for blowing and suction).

fulfills all the desired requirements. To realize an FDR device, we focused on an air jet [9]–[11]. In an air jet, airflow is ejected from a nozzle. The air jet entrains the surrounding air and allows it to be sucked. If the air jet is blocked, the air flows to the opposite side and is discharged through the suction port, which is identical to the blowing port. These suction and discharge mechanisms were incorporated into the FDR device to achieve reversal of the airflow direction with 1-DOF airflow control. The flap gate for the reversal of the airflow was produced as a part of the main body. The gate was normally closed, blocking the airflow. It was opened by the inflation of the main body caused by the air pressure of the inner flow. The amount of inflation can be controlled by the inner air pressure, which, in turn, is controlled by the input airflow rate. The closing amount of the gate could be controlled by the input airflow rate. Hence, the reversal of the airflow direction and the intensities of suction and blowing can be controlled by the input airflow rate. Another feature of an FDR device is that it is thin and flexible owing to its silicone body.

The FDR device is also effective for varying friction. Suction is useful for holding objects [6] [12] [13] because it acts as a function of adsorption, increasing surface friction. Blowing creates a thin layer of air between the FDR device and the contact target and reduces surface friction [14]. The advantage

[1]S. Nojiri is with the Graduated school of Natural science and Technology, Kanazawa University, Kakuma-machi, Kanazawa, 9201192 Japan.

[2]T. Nishimura and T. Watanabe are with the Faculty of Frontier Engineering, Institute of Science and Engineering, Kanazawa University, Kakuma-machi, Kanazawa city, Ishikawa, 9201192 Japan (e-mail: te-watanabe@ieee.org, tnishimura@se.kanazawa-u.ac.jp).

[3]K. Tadakuma is with the Graduation School of Information Sciences, Tohoku University, Sendai, Japan.

Digital Object Identifier (DOI): see top of this page.

of the thinness and flexibility of an FDR device is its adaptability. The suction function can operate on both rounded and rough surfaces. The FDR device can be attached to the rounded surfaces of robots or manipulators.

*A. Related Works*

Several methods for providing different or multiple actuating functions performed by 1-DOF fluid flow control have been developed. These methods can be categorized into two types. One method provides periodic motion [15]–[22]. Miyaki et al. developed pneumatic soft mobile robots that generated traveling waves by alternating and periodically switching the internal pressure of a chamber [15]. Tsukagoshi et al. developed an actuator that can expand and contract periodically by turning the air supply on and off [16]. Preston et al. proposed a soft ring oscillator that induces periodic motion in soft actuators with a single constant pressure source and applied it to a robot performing rolling movements [17]. Tani et al. developed an actuator that uses the deformation caused by air input to switch the flow paths and generate cyclic motion [18] [19]. Kitamura et al. proposed a soft actuator that mimicked vocal cords and generated a self-excited vibration [20]. Vasios et al. developed a soft actuator that used fluid viscosity to sequentially inflate chambers with pneumatic control of a single airflow [21]. Rothemund et al. developed a soft elastomeric valve that switched airflow paths with the membrane designed to be bistable [22]. The second type of method selectively drives a specific actuator among multiple actuators with single airflow control [7] [8] [23]. Napp et al. proposed a method of switching chambers for pressurization between multiple chambers with valves controlled by modulating the input pressure of a single airflow [7]. Nishimura et al. developed a system that switches the airflow paths for a pneumatic soft robotic hand and a friction reduction system based on the magnitude of a single-input airflow rate [8]. Ben-Haim et al. proposed switching chambers such that only the selected chamber was inflated, utilizing the viscosity of the fluid and the bistability of the elastic chambers at a single inlet flow rate [23]. The reversal of the airflow direction using a single airflow control has not been attempted. The present study addresses this issue. Note that the developed FDR device belongs to the second category, in which multiple actuators are driven by a single airflow controller.

In conventional suction grippers, such as in [12][13][24], the grasped object cannot be released simply by stopping the suction because negative pressure is maintained in the suction cup. The gripper requires solenoid valves to switch the airflow path to blow air into the suction cup for release. Thus, conventional grippers require at least one electrical cable to operate the valve, in addition to the air tube. The FDR device reduces the number of wires installed in the robotic system.

Several robotic mechanisms for varying the contact surface friction have been developed. The methods are either to increase or to reduce friction [8] [25]–[33]. Unfortunately, no attempt has been made to either increase or reduce friction using a single actuation system. This study aims to achieve both increased and decreased friction using an FDR device that can switch between blowing and suction.

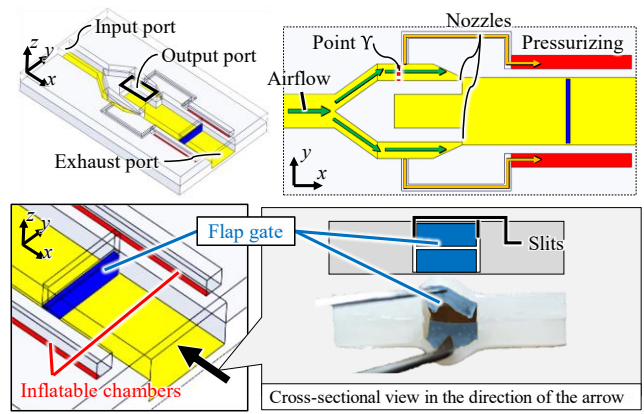

**Fig. 2.** Structure of the FDR device made of silicone (Smooth-on, Dragon Skin 10 medium).

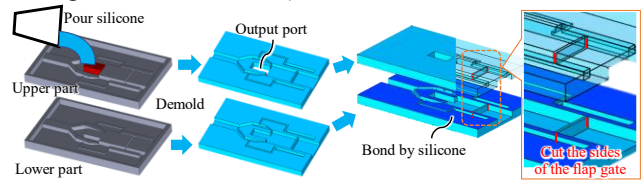

**Fig. 3.** Fabrication process of the FDR device

## II. DESIGN OF PROPOSED 1-DOF FLEXIBLE FLOW-DIRECTION REVERSAL DEVICE

The functional requirements of the FDR device are as follows: 1) flexibility under shore hardness A 10, which is close to that of human skin [34]; 2) thickness of less than 6 mm so that it can be bent by more than 90° and easily mounted on other devices such as the fingertips of a robotic hand; and 3) switching between suction and blowing realized by a single airflow control.

The structure of the FDR device is shown in Fig. 2. The FDR device has planar airflow paths. The main airflow from the input port passes through the two nozzles and is exhausted through the exhaust port. The airflow path is highlighted in yellow in Fig. 2. An output port is installed near the two nozzles for suction and blowing. The flap gate (its bottom side is highlighted in blue in Fig. 2) is installed between the nozzle exit and exhaust port. The flap gate is composed of two cantilevered walls perpendicular to the airflow path. One wall was installed on the upper side and the other on the bottom side (Fig. 2). The flap gate had the same width as the airflow path, and the two walls had slits on both sides. The walls cover the airflow path at a low airflow rate to blow the air backward, and open at a high airflow rate to create an airflow path to the exhaust port for suction. The inflatable chambers (highlighted in red in Fig. 2). are deployed on both sides of the flap gates. Air is supplied to the chambers via a flow path branching from the flow path to the nozzle. Inflatable chambers are used to open the flap gates. The entire body of the FDR device consists of silicone (Smooth-on, Dragon Skin 10 medium) with a shore hardness of A 10. Fig. 3 illustrates the fabrication process. The molds for the upper and lower parts were 3D printed (Raise 3D, Pro3). Silicone was poured into each mold. The upper and lower parts were bonded using silicone (Dragon Skin 10 medium) except

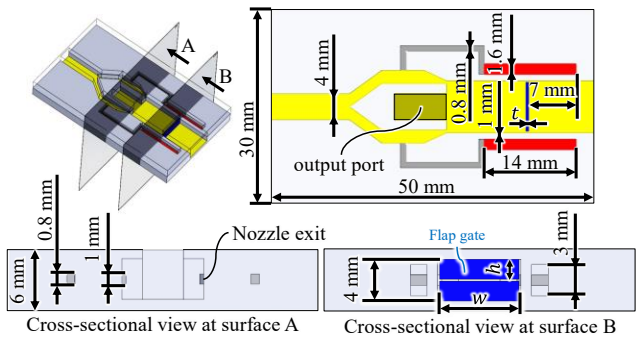

**Fig. 4.** Dimensions of the FDR device

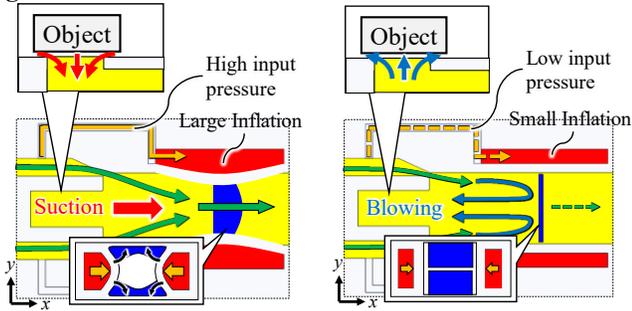

(a) Low input airflow rate  (b) High input airflow rate
**Fig. 5.** Airflow from the input port to the nozzle exit and airflow at the contact area with an object (at the output port)

for the flap gate area. The top edge of the flap gate was not coated with silicone, and the edges were kept separated. After bonding, a slit was cut between the flap gate and the inner wall of the channel. The dimensions of the FDR device are shown in Fig. 4.

The airflows from the nozzle exit to the exhaust port at high and low input airflow rates are shown in Fig. 5(a) and (b), respectively. The air passes through the nozzle, and its velocity increases according to Bernoulli's principle. The cross-sectional area of the nozzle is smallest at the nozzle exit. The air passing through the nozzle is discharged as a jet along the sidewalls of the airflow path. The downstream flow (between the nozzle outlet and exhaust port) varies with the input airflow rate, as shown in Figs. 5(a) and (b).

The amount of inflation in the inflatable chambers is determined by the input airflow rate because the air pressure in the inflatable chambers is determined by the air pressure at the air flow path (point Y) after the input flow path is bifurcated and leading to the nozzle, which is determined by the input airflow rate. The opening and closing of the flap gate were caused by inflation of the inflatable chambers. The blowing and suction in the FDR device were switched by opening and closing the flap gates.

At a low input airflow rate, the inflation of the inflatable chambers is small, and the flap gate is kept closed (Fig. 5(a)). Most of the air flows backwards, although part of the air is exhausted through a small gap in the flap gate. The air flows backward owing to the closing of the gate, and is discharged through the output port. This is the air-blowing process.

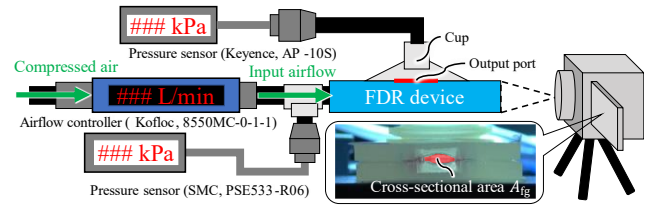

(a) Experimental setup to observe the relationship between $q_{in}$, $p_{in}$, $p_{out}$, and $A_{fg}$

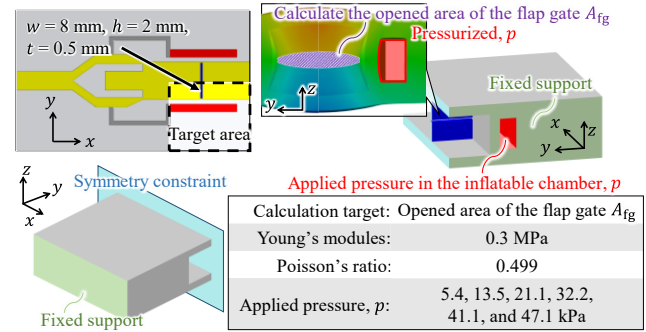

(b) Model and condition for FEM analysis

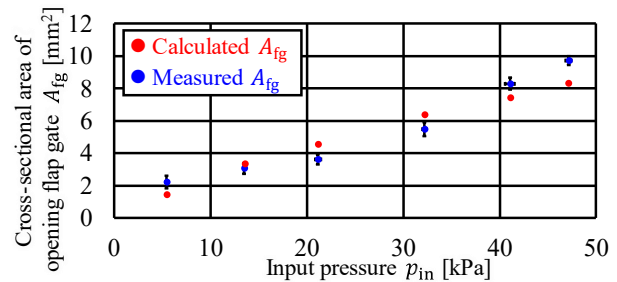

(c) Results
**Fig. 6.** Comparison of measured and calculated opening cross-sectional area of the flap gate $A_{fg}$ when varying air pressure at the input port $p_{in}$ ($\cong$ air pressure of the inflatable chamber, $p$) by varying the input airflow $q_{in}$.

A high input airflow rate causes the inflatable chambers to expand such that the walls on both sides of the flap gate protrude into the flow channel (Fig. 5(b)). With both walls protruding, the upper and lower walls of the flap gate are pressed against the upper and lower sides of the flow channel, respectively. The gate is kept open and air flows toward the exhaust port. Air is ejected from the nozzle to form an air jet. An air jet develops when air is entrained at the output port. In this process, the air at the output port is sucked. If the entrainment of the jet is disturbed, the suction capacity is reduced. In other words, the jet must flow smoothly through the exhaust port; otherwise, the suction ability is reduced. Therefore, if applying a method wherein air at a high flow rate hits the flap gate and opens it, the suction capacity is drastically reduced. The need to maintain gate elastic deformation by a high flow of air such that it remains open is another reason for this reduction. Hence, this study proposes a method that utilizes the inflation of inflatable chambers to open the gate with minimal disturbance to the airflow.

This section describes the control of the flap gate based on the input airflow rate. Based on Bernoulli's principle and continuous equation, the relationship between the air pressure $p$ at the air flow path after the input flow path (point Υ in Fig. 2) is bifurcated and the air pressure $p_{in}$ at the input port is given by

$$p = \frac{\rho}{\rho_{in}} p_{in} + \frac{\gamma - 1}{2\gamma} \rho \left(\frac{q_{in}}{A_{in}}\right)^2 \left(1 - \left(\frac{A_{in}}{2A}\right)^2\right) \quad (1)$$

where $q_{in}$ is the input airflow; $\rho_{in}$ and $A_{in}$ are the density of air and the cross-sectional area at the input port; $\rho$ and $A$ are the density of air and the cross-sectional area at the air flow path after the input flow path is bifurcated; and $\gamma$ is the specific heat ratio. Because the air pressure in the inflatable chamber is equal to the air pressure $p$ at the junction with the flow path to which the inflatable chamber is connected, (1) can express the relationship between the air pressure in the inflatable chamber and $p_{in}$. The FDR device was designed such that $A_{in} = 2A$. If the losses due to the bifurcation are assumed to be small and air density is maintained ($\rho \cong \rho_{in}$), (1) becomes

$$p \cong p_{in}. \quad (2)$$

The pressure $p$ in the inflatable chambers controls the opening and closing of the flap gate, and $p_{in}$ monotonously increases as $q_{in}$ increases. Therefore, we focused on the relationship between $p_{in}$ and the opening cross-sectional area of the flap gate, $A_{fg}$, and experimentally evaluated this relationship by varying $q_{in}$ from 5 to 30 L/min in 5 L/min increments (the corresponding $p_{in}$ = 5.4, 13.5, 21.1, 32.2, 41.1, and 47.1 kPa). Fig. 6(a) shows the experimental setup. $q_{in}$ was controlled by an airflow controller, and $p_{in}$ was measured using a pressure sensor. The experiment was conducted in triplicate under each condition. The shape of the flap gate was measured using a camera, and $A_{fg}$ was derived from the images captured using image processing. We also conducted FEM (finite element method) analysis to determine the relationship between $p$ ($\cong p_{in}$) and $A_{fg}$. Fig. 6(b) shows the model and conditions for the FEM analysis using ANSYS. The input was $p$ and the output was $A_{fg}$ in the analysis. According to [35][36], if the deformation is small (elongation of less than approximately 400%), the deformation of the silicon can be regarded as a linear elastic deformation. Therefore, the analysis was performed in the linear elastic regime. The air pressure at the airflow path near the flap gate was measured while varying $q_{in}$, with a maximum pressure of 0.7 kPa (6% of $p_{in}$). Therefore, the air pressure outside the inflatable chamber (along the airflow path, including the flap gate) was assumed to be atmospheric (0 kPa). Fig. 6(c) shows the measured and calculated $A_{fg}$ when varying $p$ ($\cong p_{in}$). The result in Fig. 6(c) shows that the measured and calculated values of $A_{fg}$ were close, and $p$ ($\cong p_{in}$) can control $A_{fg}$. Because $p_{in}$ is determined by the input airflow rate, the opening cross-sectional area of the flap gate $A_{fg}$ can be controlled by the input airflow rate.

The effects of the design parameters are discussed in the following section. Other dimensions were determined such that the process mentioned above was valid.

## III. INVESTIGATION OF DESIGN PARAMETERS

The flap gate structure is the key to switching between blowing and suction in an FDR device. Therefore, the effects of the width $w$, thickness $t$, and height $h$ of the flap gate on the air pressure (suction and blowing capacities) at the output port were experimentally investigated. The effects of nozzle exit size and material changes were also investigated.

### A. Effect of dimension of the flap gate

Table I lists the design parameters that were varied in the investigation (see Fig. 4 for the design parameters in the FDR device). The experimental setup used to measure the output airflow rate is shown in Fig. 6(a). The output port of the FDR device was completely covered with a cup. A pressure sensor was connected to the cup to measure the output air pressure $p_{out}$. We investigated $A_{fg}$ and $p_{out}$ when varying $w$, $t$, and $h$ of the flap gate. Note that varying $w$ of the flap gate indicates varying the width of the airflow channel, including the flap gate. $p_{in}$ and $p_{out}$ were measured while increasing the input airflow rate $q_{in}$ from 0 to 30 L/min at 0.1 L/min per second, while $A_{fg}$ was measured when $q_{in}$ = 5, 10, 15, 20, 25, and 30 L/min. The experiment was conducted in triplicate under each condition. $p_{out} > 0$ indicates suction and $p_{out} < 0$ indicates blowing. Note that $p_{in}$ monotonically increases as $q_{in}$ increases.

Fig. 7 shows the relationship between the input pressure $p_{in}$ and the output pressure $p_{out}$, and the relationship between $p_{in}$ and the ratio of the opening cross-sectional area of the flap gate to the cross-sectional area of the exhaust port $A_{fg}/A_{ex}$. If we focus on the case of varying $w$, the smaller the width, the larger is the positive output pressure $p_{out}$ (the larger the blowing capacity). In the case of varying $t$, the larger the thickness, the larger is the blowing capacity. A small $w$ and large $t$ are associated with a large flexural rigidity of the flap gate, because the flap gate is composed of a cantilevered wall. Thus, the larger the flexural rigidity, the larger is the blowing capacity. With respect to $h$, a lower height switches from blowing to suction with a smaller $p_{in}$. As $h$ is reduced, $A_{fg}/A_{ex}$ increases and $p_{out}$ decreases, resulting in a decrease in the $p_{in}$ at which switching occurs. The flexural rigidity of the flap gate also contributes to the switching point. The smaller the flexural rigidity of the flap gate, the smaller $p_{in}$ is for switching. Focusing on the suction state ($p_{out} < 0$), the larger the $A_{fg}/A_{ex}$ is, the smaller $p_{out}$

TABLE I  Parameters of the FDR device

| Type | Hardness (Shore A) | Nozzle size $A_{ne}$ [mm²] | Dimensions of the flap gate | | |
|---|---|---|---|---|---|
| | | | $w$ [mm] | $t$ [mm] | $h$ [mm] |
| A | 10 | 0.4 | **6** | 0.5 | 2 |
| **B*** | **10** | **0.4** | **8** | **0.5** | **2** |
| C | 10 | 0.4 | **10** | 0.5 | 2 |
| D | 10 | 0.4 | 8 | **0.4** | 2 |
| E | 10 | 0.4 | 8 | **0.6** | 2 |
| F | 10 | 0.4 | 8 | 0.5 | **1.8** |
| G | 10 | 0.4 | 8 | 0.5 | **1.9** |
| H | 10 | **0.32** | 8 | 0.5 | 2 |
| I | 10 | **0.48** | 8 | 0.5 | 2 |
| J | **20** | 0.4 | 8 | 0.5 | 2 |
| K | **30** | 0.4 | 8 | 0.5 | 2 |

*The parameter values of Type B are nominal. The values are indicated in red if they differ from the values of Type B.

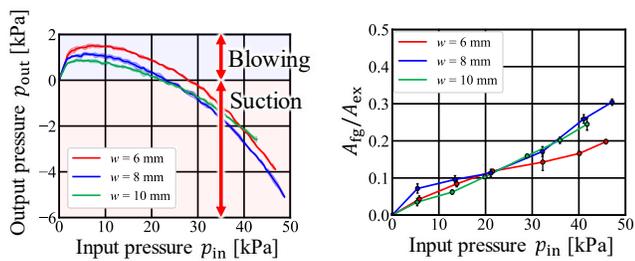

(a) When varying the width $w$ (Type A, B, and C)

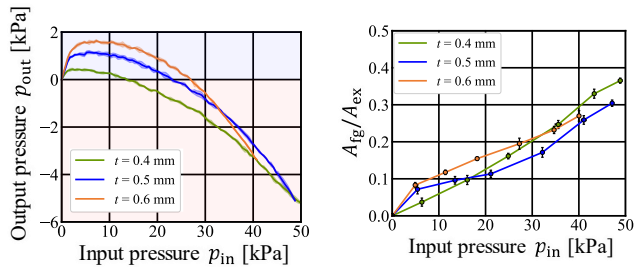

(a) When varying the thickness $t$ (Type B, D, and E)

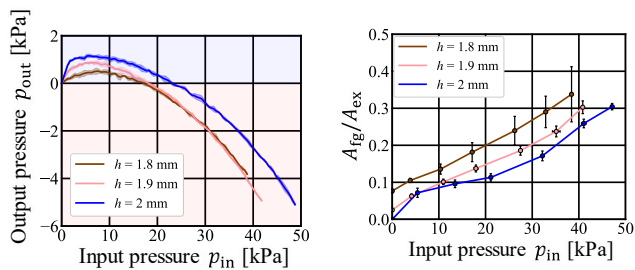

(c) When varying the height $h$ (Type B, F, and G)

**Fig. 7.** Relationship between the input pressure $p_{in}$ and output pressure $p_{out}$, and the relationship between $p_{in}$ and the ratio of the opening cross-sectional area of the flap gate to the cross-sectional area of the exhaust port $A_{fg}/A_{ex}$.

became and the greater the suction pressure in all cases. This indicates that $A_{fg}/A_{ex}$ contributes to the suction capacity; the larger the opening area of the flap gate, the larger is the suction capacity.

If $p_{in}$ is large ($p_{in} > 40$ kPa) in the case of varying $w$ (see Fig. 7(a)), $p_{out}$ for Type C ($w = 10$ mm) is larger than that for Type A ($w = 6$ mm) although $A_{fg}/A_{ex}$ for Type C was larger than that for Type A. Here, we investigated the reason for this by experimentally investigating the airflow in the flow channel, including the flap gate, during suction. The experimental setup for observing the airflow in the flow channel is illustrated in Fig. 8. The FDR device was placed on a transparent acrylic plate, such that the side without the output port was in contact with the acrylic plate to allow air to be drawn from the atmosphere. While sucking air, red droplets of water were slowly dropped into the device from the output port using a syringe. Because the body of the FDR device is transparent, the movement of the red droplets through the acrylic plate could be observed by the camera. Airflow was observed from the movement of the red droplets, that is, the direction of motion and the area where the water droplets agglomerated.

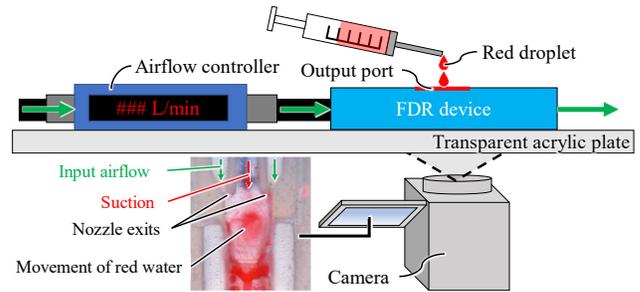

**Fig. 8.** Experimental setup for observation of inner flow.

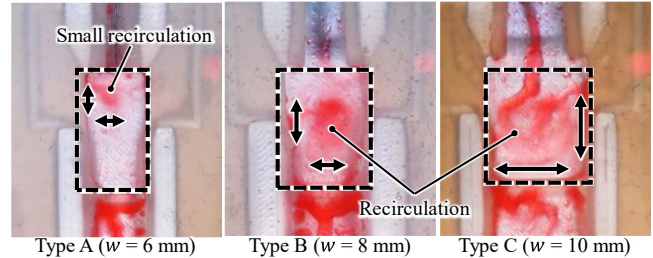

**Fig. 9.** Observed inner flow ($q_{in} = 30$ L/min).

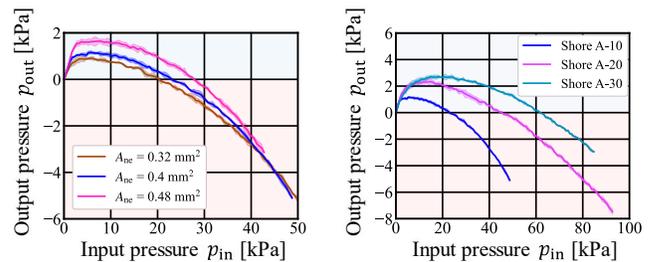

(a) When varying $A_{ne}$     (b) When varying materials

**Fig. 10.** Relationship between input pressure $p_{in}$ and output pressure $p_{out}$ when varying the cross-sectional area of the nozzle exit $A_{ne}$ and the materials.

Fig. 9 shows the observed airflows in the flow channels of Types A, B, and C (see Table I) at $q_{in} = 30$ L/min. In Types B and C, the red droplets gathered around the center of the airflow channel, indicating that the flow was reversed. Backflow can occur around jets in flow channels [9] [37] [38]. This phenomenon occurs when the jet entrains air from downstream, when the surrounding air cannot supply sufficient air for the jet to develop. This process is referred to as recirculation. Suction occurs when the jet entrains the surrounding air. If the jet entrains air from downstream, suction is not created by that air. Therefore, recirculation reduces the suction capacity. As shown in Fig. 9, the size of the recirculation area increased in the order of types A, B, and C. In the case of a recirculation area surrounded by a jet, the wider the channel, the easier it was to create a recirculation area [37]. We believe that this is why Type C had a low suction capacity. The size of the recirculation must be confirmed experimentally because of the complexity of the fluid phenomena. Remarkably, we also observed an inner flow at $q_{in} = 20$ L/min (when the suction function was available at every $w$). Similar phenomena are observed at $q_{in} = 20$ L/min.

*B. Effect of the size of the nozzle exit and material*

The effects of the nozzle-exit size and material were investigated using the experimental setup shown in Fig. 6(a).

$p_{\text{out}}$ was measured when varying the cross-sectional area of the nozzle exit $A_{\text{ne}}$, and the materials (see Table II). The experiment was conducted in triplicate under each condition. Fig. 10 shows the obtained relationship between the input pressure $p_{\text{in}}$ and output pressure $p_{\text{out}}$, while increasing the $q_{\text{in}}$ from 0 to 30 L/min at 0.1 L/min per second. As $A_{\text{ne}}$ was reduced, a larger suction capacity was obtained, and a smaller $p_{\text{in}}$ was obtained when switching from blowing to suction. This is because a smaller $A_{\text{ne}}$ results in a lower air pressure exiting the nozzle. If the rigidity of the FDR device is increased by a material change, the blowing capacity increases and $p_{\text{in}}$ required for switching also increases. This tendency is similar to that of increasing the flexural rigidity of the flap gate by varying its thickness and width.

In summary, the ratio of the opening cross-sectional area of the flap gate to the cross-sectional area of the exhaust port $A_{\text{fg}}/A_{\text{ex}}$ mainly contributes to the suction capacity, whereas the flexural rigidities of the flap gate or device itself mainly contribute to the blowing capacity. The ease of opening of the flap gate varies the point where switching from blowing to suction occurs. The aforementioned design parameters should be appropriately selected to obtain the desired suction and blowing capacities and switching points.

## IV. FRICTION VARIATION FUNCTION OF THE FDR DEVICE

The air blowing and suction functions of the FDR device can contribute to the friction variation. In this study, the friction variation function of an FDR device was evaluated. Frictional forces were measured to evaluate the maximum static and kinetic friction coefficients when an object slid on the FDR device.

Fig. 11(a) shows the experimental setup used to measure the frictional forces. The FDR device was fixed on a table with the output port facing upward. We used a weight composed of fabric and a water bag to completely cover the output port of the FDR device and ensure that the pressure applied to the entire contact surface was as uniform as possible. Fabrics are generally difficult to suction and are useful for evaluating suction capacity; thus, fabric was selected as the target object or contact surface. The water bag and fabric were clipped to allow them to move together. We investigated cases with total weight of 16.0 gf (= 0.157 N), 100 gf (= 0.981 N), and 200 gf (= 1.96 N). A load cell was attached to the clip using a string. The position of the load cell was controlled using a linear actuator (IMADA, RF34) to move the weight of the FDR device. The velocity of the weight was set to 1 mm/s. The frictional force was measured as the force applied to the load cell while pulling the weight. A type B FDR device (see Table I) was used. The input airflow rates $q_{\text{in}}$ were set to 0, 10 (blowing), 20 (suction), and 30 L/min (suction) using an airflow controller (KOFLOC, 8550MC-0-1-1). The experiment was conducted three times for each $q_{\text{in}}$. The coefficients of the maximum static friction $\mu_{\text{s}}$ and kinetic friction $\mu_{\text{k}}$ are derived as follows:

$$\mu_{\text{s}} = f_{\text{slip}}/W, \quad \mu_{\text{k}} = \bar{f}/W. \tag{3}$$

where $W$ is the load by weight (0.157, 0.981, or 1.96 N), $f_{\text{slip}}$ is the friction force when slippage occurs (detected by video),

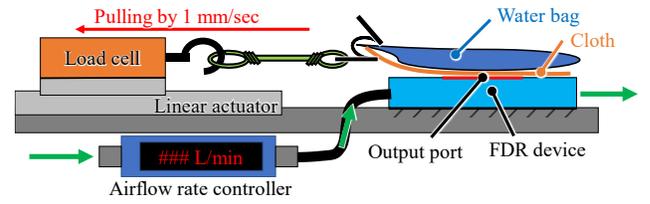

(a) Experimental setup.

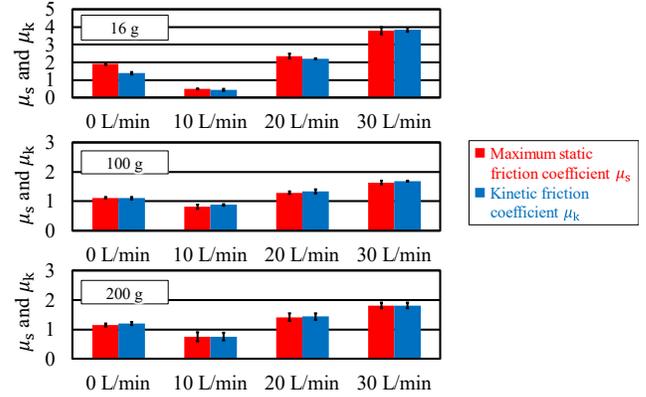

(b) Coefficient of friction for each input airflow rate.

**Fig. 11.** Friction variation of the FDR device.

and $\bar{f}$ is the mean friction force when the weight moves on the FDR device. Fig. 11(b) shows the mean values of $\mu_{\text{s}}$ and $\mu_{\text{k}}$ for each $q_{\text{in}}$. Regardless of the weight of the object, the blowing mode ($q_{in}$ = 10 L/min) decreased the coefficient of friction and the suction mode ($q_{\text{in}}$ = 20 L/min and 30 L/min) increased it. Larger suction capacity yielded larger $\mu_{\text{s}}$ and $\mu_{\text{k}}$. The suction function increases the normal force, thereby contributing to a greater frictional force, whereas the blowing function decreases the normal force, thereby contributing to a smaller frictional force. This is the mechanism of variable friction in the FDR device (Fig. 5). Therefore, the rate of increase in the coefficient of friction decreased as the normal force (weight of the object) increased (see Fig. 11(b)). The results demonstrated that the FDR device could achieve both increased and decreased friction with a single airflow control. The results also demonstrate that the FDR device can be sucked into the fabric.

## V. EXPERIMENTAL APPLICATION OF THE FDR DEVICE

In this section, the air-blowing and suction functions of the FDR device were investigated by simulating tasks.

### A. Picking up a thin object covered with dust.

Blowing and suction were used to remove lightweight obstacles and pick up objects, respectively. The purpose of this experiment was to demonstrate that an FDR device can be used to remove lightweight obstacles (dust) and pick up objects. Another purpose of this experiment was to validate blowing to release an object when it sticks to the soft surface of the FDR device. The experimental setup is shown in Fig. 12(a). The FDR device was fixed to an automatic positioning stage (IMADA, RF34) at 45 °. The target object was placed on a conductive rubber sheet to prevent the generation of static electricity. The objects were a polypropylene (PP) sheet (0.25 mm thickness), a piece of paper (0.2 mm thickness), and a nonwoven fabric sheet (0.3 mm thickness) with a size of 30 × 30 mm. The obstacles were small pieces of nylon thread that were scattered over the object. The experimental procedure was as follows. 1) The input airflow was set to 10 L/min to start blowing. 2) The FDR device was moved to an object that flowed away from the obstacles. The motion stopped when the FDR device made contact with the edge of the object. 3)The input airflow was set to 30 L/min to start suction and pick up the object. The FDR device was then moved upwards. Finally, the input airflow was stopped. If the object was not released, the input airflow was set to 10 L/min, and the blowing started again. Fig. 12(b) shows a representative result of picking up a PP sheet. Both obstacles were removed by blowing and picking up all the objects by suction. Sticking of the FDR device after picking up was observed on the PP sheets. In this case, the PP sheet was released by blowing. The results demonstrate the effectiveness of the FDR device.

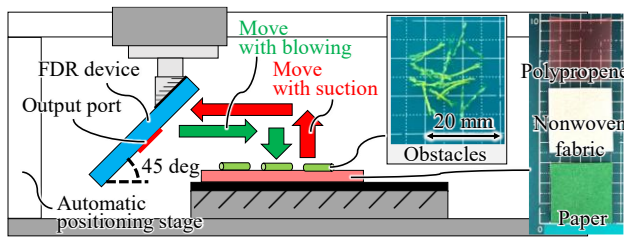

(a) Experimental setup.

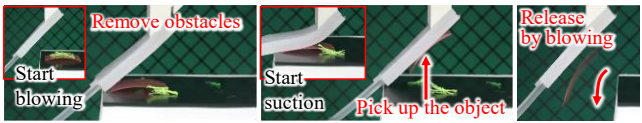

(b) Representative results for the PP sheet

**Fig. 12.** Schematic of removal obstacles and picking up an object.

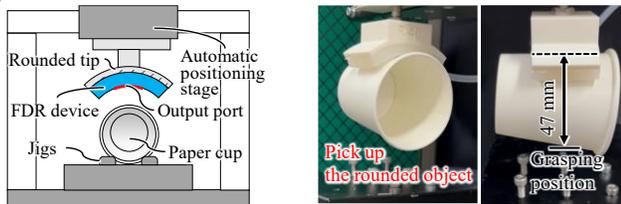

(a) Experimental setup  (b) Result

**Fig. 13.** Paper cup grasping test when the FDR device is attached to a rounded surface.

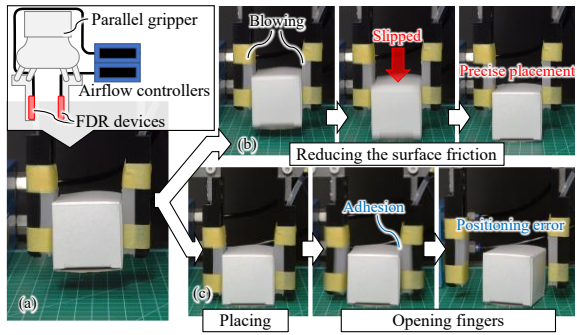

**Fig. 14.** Grasping and releasing a paper box with a gripper equipped with an FDR device on each finger surface: (a) experimental setup, (b) release by blowing, and (c) release by opening the fingers.

We also conducted a grasping test on a paper cup, while the FDR device was attached to a rounded tip attached to an automatic positioning stage (IMADA, RF34). The radius of the rounded tip was 40 mm, the diameter of the paper cup at the grasping position was 47 mm, and its weight was 2.4 g. As shown in Fig. 13, the results demonstrate that the flexibility and thinness of the FDR enable the device to operate on a rounded surface and render it effective in picking up a rounded object.

### B. Use as a finger surface

The FDR device can be used as a surface for robotic fingers to facilitate the grasping and release of an object with the suction and blowing functions. The thinness and softness of the FDR device facilitate its attachment to robotic fingers. The FDR device attached to a finger can function as a variable-friction system on the finger surface, as described in Section IV. The increased friction due to the suction function increases the grasping stability. The gradual decrease in frictional force through air pressure control enables the release of an object without changing the width of the finger opening. This eliminates the uncertainties caused by the opening motion of the fingers and adherence of the object to the finger surface and ensures accurate placement while maintaining the grasping posture. The soft finger surface conforms to the shape of the object and enhances grasping stability, whereas its large friction makes release of the grasped object difficult [8]. This function of accurate placement is particularly effective when the finger surface is soft. To demonstrate its effectiveness, we conducted an experiment to grasp and release an object (a paper box, weight 22 g) using a two-finger gripper [39] equipped with an FDR device on each finger surface. The two FDR devices were connected to separate airflow controllers. First, the gripper picked up the object using two fingers (Fig. 14 (a)). Subsequently, the object was released without changing the width of the finger opening while changing the airflow from positive to negative pressure (Fig. 14(b)). For comparison, we conducted an experiment involving picking up and placing an object without blowing (Fig. 14(c)). In this case, the position of the object was disturbed when the fingers opened for release. The results demonstrate the effectiveness of the FDR device in facilitating stable grasping and accurate placement with a variable friction function.

## VI. CONCLUSIONS

In this study, we proposed an FDR device, which is a slim and flexible soft robotic device capable of switching between airflow and suction with a single airflow control. The FDR device achieved suction by entraining the surrounding air by the nozzle jet stream and blowing by reversing the airflow through the flap gate. Blowing was activated when the flap gate was closed, whereas suction was activated when the flap gate opened. The opening and closing of the flap gate were controlled by the expansion of inflatable chambers installed near the gate. The extent of the expansion was determined by the pressure of the input airflow. Switching between air blowing and suction was achieved by controlling the input airflow rate. All the components were made of a silicone body to achieve thinness and flexibility. We introduced design parameters including the width, thickness, and height of the flap gate, and demonstrated that it contributed to the blowing and suction capacities. We demonstrated that the FDR device can function as a variable friction system by using airflow and suction, and the function is effective for facilitating stable grasping and accurate placement. We also demonstrated that the FDR device is effective in picking up objects covered with lightweight obstacles. The device removed obstacles by blowing and picked up objects through suction. Blowing has also been proven to be effective in releasing objects. Another experiment demonstrated that the flexibility and thinness of the FDR device enabled its operation on a rounded surface. In the future, we plan to expand the range of applications by miniaturizing the FDR device; optimizing the design parameters according to target operations, including energy efficiency; and developing robotic application systems, such as object manipulation and human assistance systems, using the proposed FDR device.